# Ontology-Driven Self-Supervision for Adverse Childhood Experiences Identification Using Social Media Datasets

Jinge Wu[1,2], Rowena Smith[2] and Honghan Wu[1]
[1]*Institute of Health Informatics, University College London, London, UK*
[2]*Usher Institute, University of Edinburgh, Edinburgh, UK*
*{jinge.wu.20, honghan.wu}@ucl.ac.uk, rjss18@gmail.com*

Abstract: Adverse Childhood Experiences (ACEs) are defined as a collection of highly stressful, and potentially traumatic, events or circumstances that occur throughout childhood and/or adolescence. They have been shown to be associated with increased risks of mental health diseases or other abnormal behaviours in later lives. However, the identification of ACEs from textual data with Natural Language Processing (NLP) is challenging because (a) there are no NLP ready ACE ontologies; (b) there are few resources available for machine learning, necessitating the data annotation from clinical experts; (c) costly annotations by domain experts and large number of documents for supporting large machine learning models. In this paper, we present an ontology-driven self-supervised approach (derive concept embeddings using an auto-encoder from baseline NLP results) for producing a publicly available resource that would support large-scale machine learning (e.g., training transformer based large language models) on social media corpus. This resource as well as the proposed approach are aimed to facilitate the community in training transferable NLP models for effectively surfacing ACEs in low-resource scenarios like NLP on clinical notes within Electronic Health Records. The resource including a list of ACE ontology terms, ACE concept embeddings and the NLP annotated corpus is available at https://github.com/knowlab/ACE-NLP.

## 1 INTRODUCTION

Adverse childhood experiences (ACEs) encompass a broad variety of early traumas that occur in childhood, including both direct (e.g., abuse and neglect) and indirect (e.g., family mental illness and domestic violence) types (Hughes et al., 2017; Felitti et al., 1998). Numerous studies have found that ACEs are consistently and positively associated with high rates of severe adverse events, such as serious mental disorders (suicide attempts, major depression, substance abuse, adult victimization, etc.) (Liu et al., 2021). Moreover, the lifetime frequency of ACEs is significantly greater in homeless individuals than in the general population, indicating ACE exposure may be related to an increased risk of mental illness, drug addiction, and victimization (Liu et al., 2021). Policy actions and evidence-based interventions are urgently required to prevent ACE and to address the unfavourable consequences linked with this group of people.

In order to promote the surveillance and study of adverse childhood experiences (ACEs), Brenas et al. standardized the ACE ontology and has been made publicly available as a formal reusable resource that can be utilised by the mental health research (Nagowah et al., 2021; Brenas et al., 2019). The state-of-the-art ACEs Ontology (ACESO) has been available to the mental health community as well as the general public through BioPortal. Further in 2021, Ammar et al. unveiled Semantic Platform for Adverse Childhood Experiences Surveillance (SPACES), a novel and explainable multimodal AI platform to assist ACEs surveillance, diagnosis of associated health issues, and subsequent therapies (Ammar et al., 2021). Through the product, they generate suggestions and insights for better resource allocation and care management.

However, existing ontologies are not directly Natural Language Processing (NLP) applicable. There are at least two issues. First, ACE ontologies cover both high-level concepts (e.g., Disease or Findings) for grouping concepts and more specific

ACE concepts like child abuse. Unfortunately, existing ontologies do not provide information on which concepts are ACEs specifically, making it impossible to use the ontologies directly for NLP. Second, only a subset of ACE ontology concepts are mapped to NLP-friendly terminologies like SNOMED-CT. For those concepts introduced in these ontologies specifically there are no definitions of synonyms. This makes them less useful for NLP tasks like named entity recognition. Aiming to close these gaps, this study aims to make an improvement on the previous ACE ontologies by refining ACE concepts and their corresponding mappings to Unified Medical Language System (UMLS), which is much more NLP friendly and has been widely used in clinical NLP community.

Another challenge for ACE identification with NLP is the lack of annotated data. Annotating a variety of clinical notes from scratch requires specific domain expertise and also research access to a large corpus (to have enough cases). Both of these are costly, in many cases not feasible, when dealing with sensible electronic health records (EHRs), impeding the development of effective automated approaches for ACE identification from EHRs. This study proposes a practical and effective approach leveraging ontologies and self-supervision to alleviate the burden of annotation.

To address the above-mentioned two challenges, this paper proposes a refined ACE terminology with an off-the-shelf NLP tool to conduct a preliminary study on one open accessible corpus (Reddit) for identifying ACEs. Specifically, we choose SemEHR, an open-source toolkit for identifying mentions of UMLS concepts from textual data. Then, we propose a self-supervision approach on document-concept matrix to derive concept embeddings for facilitating document classification (i.e., identify types of ACE documents). Our experiments show that without any supervision, our approach can achieve reasonable performances both on named entity recognition and document classification. This preliminary study will provide valuable experiences, providing reusable resources for the community and identifying future research directions for our follow-up studies, aiming for developing efficient tools and resources for supporting the automated extraction of ACEs from clinical notes within EHRs.

## 2 METHODOLOGY

### 2.1 Data Collection from Reddit

Instead of clinical notes, this paper considers publicly available social media corpus since it is easier to access and potentially large scale for training large models. We obtain dataset from Reddit (Low et al., 2020), containing 32439 posts and text features between 2018 and 2020 on the topic of mental health.

### 2.2 ACE-Defined Terms

The public ACE ontology, which is ACESO, covers a wide range of ACEs containing 297 classes, 93 object properties, and 3 data properties with external links to SNOMED-CT and other ontology sources. However, there remain problems with ontology mappings, which lack consistent external links to standard clinical terminology systems. To solve this problem, we propose an approach by linking the ACE concepts to the widely used Unified Medical Language System (UMLS). Finding the leaf nodes from the ontology is an efficient way to obtain narrower ACE concepts. Therefore, we extract the leaf nodes from the ACESO and further construct one-to-one mapping to UMLS concepts. Moreover, given that the dataset utilised in this paper is from social media, it is less likely to identify sufficient childhood adverse events. As a result, we extend our concepts to broader adverse events in mental health. To be more specific, all concepts of mental disorders (including all descendants) and their adverse events are included.

### 2.3 Identifying ACEs on Reddit Using NLP Tools

This paper applied SemEHR to find ACE mentions in text files with the terms/concepts mentioned above (Wu et al., 2018). It first conducted Named entity recognition (NER) of all UMLS concepts (with their UMLS CUIs) from the texts and filtered ACE mentions based on the concept lists mentioned above. After that, some Information extraction (IE) rules are applied to filter unwanted mentions. It also associated semantic types of annotations and their clinical contexts with dedicated extraction rules, allowing for improved IE capabilities like generating structured vital sign data from observation notes.

### 2.4 Self-Supervision by Concept Embeddings

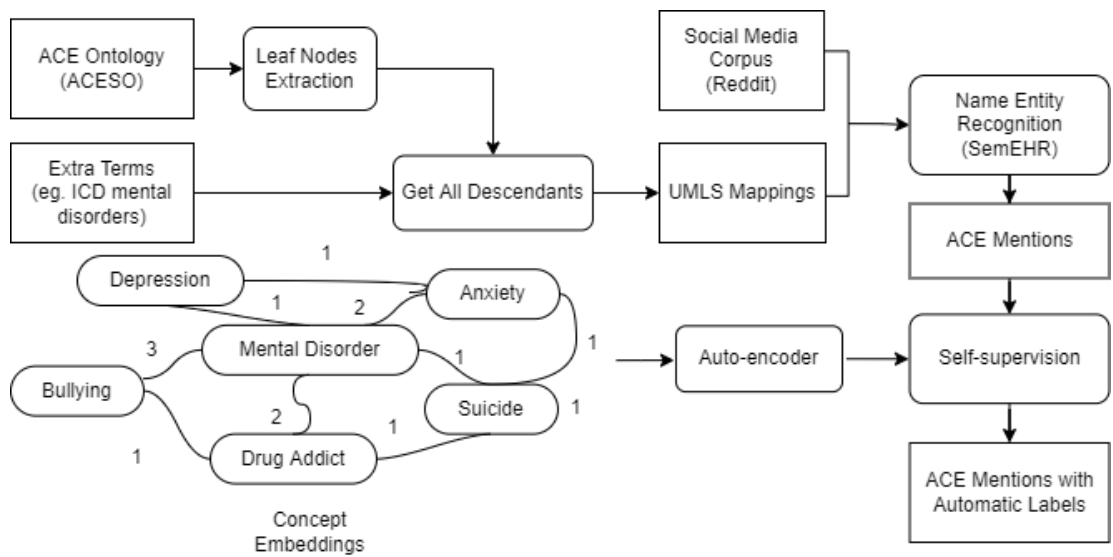

Figure 1. This is the flowchart of our work. To specify, we will combine ACESO and extra terms/concepts by finding the leaf nodes from ACESO and getting all the descendants of both term lists. SemEHR will look for relative mentions from Reddit based on these mapped concepts from UMLS. Self-supervision with graph links will be designed based on the outputs from SemEHR. This will automatically provide labels for the dataset.

This paper then considers a self-supervision-based approach to help improve the performance of ACE identification. We developed a so-called "concept embeddings", which is a concept graph with the co-occurrence of all individual terms and their co-occurrence appear in each sample (document). To be more specific, the concept graph is constructed by two matrices: concept-level matrix and document-level matrix. The concept-level matrix is an $m \times m$ square matrix ($m$ is the number of unique terms found in all documents), describing the co-occurrence of each concept. The document-level matrix is in the size of $n \times m$, where $n$ indicates the number of documents. This matrix measures how terms are co-related in each document. The aim for concept embeddings is to take the contextual information into consideration. Then cosine similarity is calculated for the two matrices at concept level and document level. We selected various thresholds to automatically label the mentions and calculated the precision and recall.

### 2.5 Auto-Encoder for Dimensionality Reduction

One issue remaining is that the concept embeddings we obtained are sparse matrices (i.e., matrices that are comprised of mostly zero values). This will lead to low efficiency when running large-scale models. To solve this, this paper considers the auto-encoder as a tool for dimensionality reduction. Auto-encoder is an unsupervised Artificial Neural Network that attempts to encode the data by compressing it into the lower dimensions and then decoding the data to reconstruct the original input. In this case, it is feasible to train an auto-encoder and compress the concept/document vector to a lower dimension for computing the cosine similarity.

## 3 EXPERIMENTS AND RESULTS

To begin with, we designed a list of 1715 concepts in total, including leaf nodes from ACESO and mental disorders with all descendant concepts. After that, the NLP tool, SemEHR is applied with Named entity recognition (NER) task on these 1715 concepts. As a result, 322 out of 1715 unique concepts are found from datasets. Due to the limitations of manual annotation, a random choice of 50 samples is selected for annotation by domain experts. We considered SemEHR as the baseline model of this classification task. It achieved 85.3 of precision, 70.7 of recall and 77.3 of F1-score. The score of precision is slightly higher than recall, indicating that it operates well on positive samples.

We then developed self-supervision with concept embeddings using the 322 terms and trained auto-encoder on concept embeddings to reduce the dimension size from 322 to 50. Figure 2 shows the precision-recall area under curve (AUC) for different thresholds with and without using auto-encoder. In general, the precision-recall AUC illustrates the trade-off between precision and recall, where high recall relates to a low false negative rate, and high precision relates to a low false positive rate.

As can be seen from Figure 2, the precision-recall AUC seems to be in the same pattern for those with auto-encoder and without auto-encoder. This indicates our pretrained auto-encoder retains almost all useful information from the original vectors.

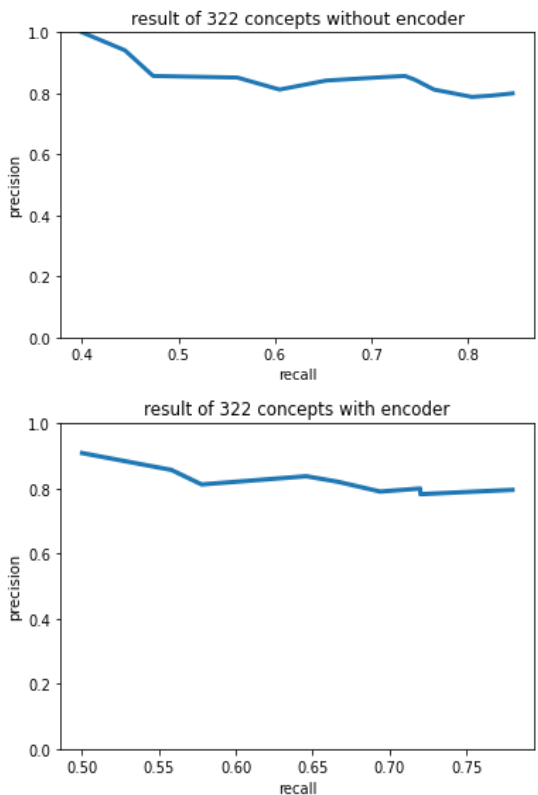

Figure 2. Precision-recall area under curve (AUC) for different thresholds with and without auto-encoder. For the auto-encoder, we compressed the 322 concepts vector into a 50-dimension vector.

We also consider the performance for each concept. There are 26 concepts identified from the 50 Reddit annotated datasets and the recall, precision and F1-score are calculated for each concept. Figure 3 shows the most frequent 5 concepts and an average performance of all other mentioned mental disorders. In general, the model gives high score on both precision and recall.

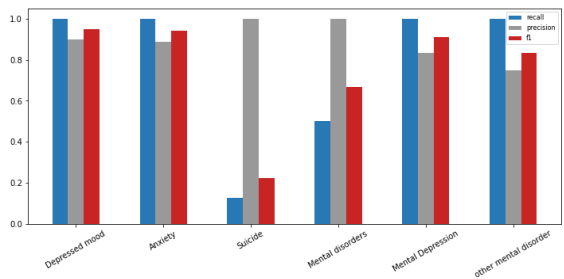

Figure 3. Performance of individual concepts. It is noticeable that the term "suicide" gives a high score in precision but low in recall. The reason for this could be that the true positives for this term is low compared with the ground truth. In other words, there are many terms to express suicidal thoughts, for example, "end my life", however, the model failed to identify this.

## 4   DISCUSSIONS

To start with, this paper contributed to an enhanced ACE ontology for natural language processing, with more comprehensive concepts and consistent mappings to UMLS. Our results have shown that the advanced ACE concepts make significant improvements in identifying ACE mentions from the social media corpus. However, we noticed that the dataset used in this paper is from social media, which might have different language, structure and even biased samples compared to EHR datasets. For example, when people express suicidal thoughts, they are more likely to use "kill myself" instead of the exact word from UMLS. To solve this, we try to include as many similar terms as possible to fix it. And the extra terms are mapped to their top concepts so that they can be learned with relations. Moreover, we consider the concept embeddings trained from EHR data as the next step for a further implementation.

In addition, this paper proposed self-supervision with the technology of concept embeddings by representing the co-occurrence of concepts. This approach has its strength in finding contextual information for classification and automatic labelling. Another point is that, based on the annotation results, it is noticeable that one word can be treated differently in different scenarios. For instance, there might be situations where 1 of the 5 "anxiety" mentions is a false positive based on its contextual environment. In this case, it is not likely to be identified by self-supervision using the current graphs. To improve its performance, future work could be done by distributing weighted frequency on the same concepts in each document.

Furthermore, this paper introduced auto-encoder as a tool for dimensionality reduction. The pre-trained auto-encoder can compress a large scale matrix into a two-dimensional matrix. This is of vital importance when training large sparse matrix with machine learning algorithms. Our results show good fit on the training embeddings by retaining as much information retention of the original data as possible.

# 5 CONCLUSIONS

In summary, this paper described an effort to use knowledge representation, natural language processing and machine learning for compiling a suite of resources for facilitating automated ACE identification from free-text data particularly in low-resource environments (i.e., labelled data is scarce). In particular, based on the previous ACE terms, this paper identified extra ontological terms from UMLS and developed comprehensive mappings to its narrower concepts. We also proposed practical approaches to self-supervision by utilising the concept co-occurrence embeddings. This will enable us to automatically label the textual data, leading to minimal manual annotations for training supervised learning models such as deep learning NLP models. Furthermore, to create a compact representation of reusable concept embeddings, we trained an auto-encoder model to reduce the dimension of the sparse concept-document matrix. The model can compress useful information into matrix with much fewer dimensions, leading to efficient computation for utilising semantics of concept embeddings. In addition, this paper identified ACEs from Reddit, leading to publicly available resources for benchmarking.

This work is very much working in progress. There are many areas that need further developments and improvements. Specifically, future steps would involve 1) further exploration with EHR datasets and combine the concept embedding from social media and EHR together; 2) enhancement on the current concept representation with more attention on the same concept; and 3) create a public accessible benchmark on Reddit datasets and other publicly available datasets (such as tweets) for ACE identification with gold-standard annotations by domain experts.

# ACKNOWLEDGEMENTS

This research is supported by the UK's National Institute for Health Research (grant number: NIHR202639) and the UK's Medical Research Council (grant number: MR/S004149/2).

# APPENDIX

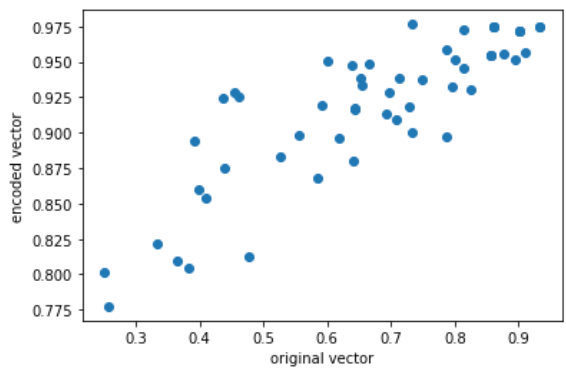

Figure 4. This is an example of annotations. The words highlighted in blue are those truly identified by NLP (i.e. true positive). The words in green are manually annotated as either "Not_ACEs" or "Manual_ACEs". "Not_ACEs" refers to those false terms identified by NLP (i.e. false positive). "Manual_ACEs" refers to those true terms not identified by NLP (i.e. false negative).

Figure 5. This is the scatter plot of original vector vs encoded vector. As can be seen, it shows patterns of linear relations among them, indicating the encoded vector has retained the majority information of the original vector.

Equation 1 shows cosine similarity mathematically. If we consider A and B as two vectors, the cosine similarity is defined as the cosine of the angle between them.

$$\cos(\theta) = \frac{A \cdot B}{\|A\|\|B\|} = \frac{\sum_{i=1}^{n} A_i B_i}{\sqrt{\sum_{i=1}^{n} A_i^2} \sqrt{\sum_{i=1}^{n} B_i^2}} \quad (1)$$

The cosine similarity is scaled within the interval [-1,1]. The value is close to 1 means the cosine of the angle is small, which means that two vectors are similar. The value is 0 indicating orthogonality. The value is close to -1 meaning approximate the opposite.